\documentclass[letterpaper]{article}

\usepackage[letterpaper]{geometry}

\usepackage[utf8]{inputenc}
\usepackage[T1]{fontenc}
\usepackage[cjk]{kotex}

\usepackage{amsmath}
\usepackage{graphicx}
\usepackage{url}
\usepackage{xcolor}
\usepackage{latexsym}

\usepackage{hyperref}
\definecolor{darkblue}{rgb}{0, 0, 0.5}
\hypersetup{colorlinks=true,citecolor=darkblue, linkcolor=darkblue, urlcolor=darkblue}
\usepackage{tabularx}
\usepackage{datetime}

\usepackage{setspace}
\usepackage{lineno}

\usepackage{longtable}
\usepackage{booktabs}

\usepackage{csquotes}
\usepackage{algorithmic,algorithm}
\renewcommand{\algorithmiccomment}[1]{\bgroup\hfill$\triangleright$~#1\egroup}

\usepackage{hyperref}
\usepackage[authoryear,round]{natbib}

\usepackage[linguistics]{forest}
\usepackage{synttree}
\usepackage{subcaption}
\usepackage{tikz-dependency}
\usepackage{tikz}
\usetikzlibrary{positioning,calc}

\usepackage{comment}
\usepackage{float}

\newcommand{\zh}[1]{\begin{CJK}{UTF8}{gbsn}#1\end{CJK}}
\newcommand{\zhgloss}[3]{\zh{#1} \textit{#2} (`#3')}
\newcommand{\errcorr}[6]{*\zhgloss{#1}{#2}{#3} $\rightarrow$ \zhgloss{#4}{#5}{#6}}

\usepackage{langsci-gb4e}
\usepackage{booktabs,tabularx}
\newcolumntype{Y}{>{\raggedright\arraybackslash}X}

\title{\textbf{\textit{A Layered Taxonomy for Chinese Learner Grammatical Error Annotation}}}
\author{
\begin{tabular}{ccc}
 Mengyang Qiu     &&  Jungyeul Park\\
Saint Elizabeth University    &&  KAIST \\
USA    &&  South Korea \\
{\tt mqiu@steu.edu} 
&$\qquad$& {\tt jungyeul@kaist.ac.kr} 
\end{tabular}
}
\date{
}

\begin{document}

\maketitle

\begin{abstract}
Grammatical error annotation in Chinese learner writing requires labels that are both consistent and linguistically meaningful. This paper proposes a layered scheme linking computational Chinese grammatical error correction (CGEC) with pedagogical error analysis. The scheme first identifies character- and punctuation-level orthographic errors, labeling them by edit operation and subtype. Other errors receive a three-layer core label combining edit operation, linguistic domain, and part of speech, with optional Chinese-specific extensions for aspect, modality, comparison, argument structure, and complements. Drawing on CGEC resources, learner-error taxonomies, and Mandarin grammar, the taxonomy is evaluated through a coverage analysis of automatically extracted MuCGEC edits and a preliminary consistency study in which five large language models apply it to a sample. The results support the layered approach while identifying category boundaries requiring further refinement.
\end{abstract}


\doublespacing
\section{Introduction}

Grammatical error annotation is a foundational activity in the study of second-language writing. By classifying the errors that learners produce, researchers can examine trajectories of interlanguage development and language-specific learning challenges, teachers can provide targeted formative feedback, and developers can evaluate writing-assistance tools. The usefulness of these annotations depends on the taxonomy underlying them. Categories that are too broad conceal meaningful differences, whereas categories that are too detailed may be difficult to apply consistently. An effective taxonomy must therefore balance descriptive precision with practical usability \citep{ludeling-hirschmann-2015-error,eryiugit-etal-2025-error}.

Chinese presents a particularly instructive case.\footnote{Chinese here refers specifically to Modern Standard Mandarin, the target language of the reviewed corpora and L2 curricula.} In the Chinese writing system, characters typically associate a written form with a syllable and a morpheme, creating close relationships among visual form, sound, and meaning. Written Chinese does not mark word boundaries with spaces, so annotators and automatic tools may segment the same sentence differently. Moreover, as a language with little inflectional morphology, Mandarin makes relatively limited use of changes in word form to mark tense, number, or case. Grammatical meaning instead depends heavily on word order, aspect markers, structural particles, and particular constructions \citep{li-thompson-1981-mandarin,packard-2000-morphology}. Chinese learner writing consequently includes homophone confusions associated with pinyin input, substitutions among visually similar characters, misuse of the three \textit{de} particles (\zh{的} \textit{de}, \zh{地} \textit{de}, and \zh{得} \textit{de}), and errors in constructions such as \zh{把} \textit{bǎ} and \zh{被} \textit{bèi} \citep{dazhong-2020-analysis,gu-etal-2025-improving}.

Computational Chinese grammatical error correction (CGEC) annotation commonly begins with a learner sentence and one or more reference corrections. An automatic program aligns the versions, identifies the edited spans, and classifies each edit according to its surface operation and relevant linguistic properties. ERRANT established this approach for English \citep{bryant-etal-2017-automatic}; ChERRANT adapted it to Chinese for the Multi-reference Chinese Grammatical Error Correction (MuCGEC) benchmark dataset \citep{zhang-etal-2022-mucgec}; and \citeauthor{gu-etal-2025-improving}'s (\citeyear{gu-etal-2025-improving}) Chinese implementation added distinctions based on character sound, character shape, the \textit{de} particles, and types of reordering. These tools facilitate the annotation and comparison of large datasets, particularly for system evaluation \citep{qiu-etal-2025-chinese}.

Learner-corpus research and Chinese pedagogical grammar approach learner errors from a different perspective. Their classifications describe recurrent learning difficulties, including the character--vocabulary--sentence hierarchy of \citet{linlin-2006-error} and the detailed functional and constructional categories of \citet{dazhong-2020-analysis}.

The computational and pedagogical traditions thus leave a practical gap. Computational schemes provide stable labels such as \texttt{R:VERB}, but say relatively little about the grammatical contrast a learner needs to understand. Pedagogical classifications provide richer diagnoses, but often lack a consistent representation for corpus annotation. To bridge this gap, the present study develops a layered taxonomy that preserves the reproducibility of edit-based computational annotation while accommodating the functional and constructional distinctions needed for pedagogical diagnosis. In doing so, it follows general recommendations for error annotation by recording the edited unit, surface change, linguistic level, and relevant metadata as separate dimensions \citep{eryiugit-etal-2025-error}.

The proposed taxonomy organizes error information into distinct, modular layers. Every error is grounded first in an observable \textbf{edit operation} (Missing, Replacement, Unnecessary, Word Order, or Character Order). An initial screening pass identifies character- and punctuation-level orthographic errors, assigning them an operation-and-subtype label of the form \textsc{op:orth:subtype}, such as \texttt{R:ORTH:phon} or \texttt{CO:ORTH:order}, without a part-of-speech tag or pedagogical extension. All remaining errors receive a three-layer core label (\textsc{op:dom:pos}) combining the edit operation, a broad \textbf{linguistic domain} (Lexical-Content, Lexical-Functional, or Structural), and a \textbf{part-of-speech} tag. An optional extension layer can add finer-grained functional and constructional categories to these non-orthographic labels, including aspect, modality, argument structure, and complement formation. This design allows projects to share compact, standardized labels while adding diagnostic detail according to their pedagogical aims.

The explicit specification is intended to support LLM-assisted annotation as well as manual annotation and conventional automatic tools. Recent learner-corpus studies have used large language models (LLMs) to propose error locations, corrections, and taxonomic labels for human review \citep{gajo-etal-2025-learn,acharya-etal-2025-tracing}, while Chinese GEC research has used them to generate explanations and evaluate corrections \citep{li-etal-2025-rethinking}. These applications make clear definitions and decision rules important for models as well as human annotators.

This study makes four contributions. First, it synthesizes design insights from influential CGEC resources and pedagogical taxonomies of Chinese learner errors. Second, it integrates these insights into a two-route annotation architecture, with optional pedagogical extensions for non-orthographic errors. Third, it provides a comprehensive inventory of category definitions, decision rules, and worked examples that projects can adopt modularly according to their needs. Finally, it evaluates the proposed taxonomy through a category-coverage analysis of automatically extracted edits from the MuCGEC benchmark and a preliminary investigation of annotation consistency across five LLMs.

The remainder of this paper is organized as follows. Section~\ref{sec:existing} reviews prior resources and describes the taxonomy design procedure. Sections~\ref{sec:taxonomy}, \ref{sec:extensions}, and \ref{sec:comprehensive} present the annotation scheme, pedagogical extensions, and reference specification, respectively. Section~\ref{sec:adequacy} evaluates the taxonomy, Section~\ref{sec:discussion} discusses its implications and limitations, and Section~\ref{sec:conclusion} concludes. The online supplement provides detailed comparisons with earlier resources, expanded pedagogical tables, worked mappings, and the complete protocol and additional results for the LLM study.\footnote{Anonymized supplementary materials are available at \url{https://anonymous.4open.science/r/new-cgec-taxonomy-CAFC/}.}

\section{Prior resources, taxonomies, and design procedure}\label{sec:existing}

This section reviews the datasets, tools, and taxonomies that motivate the layered scheme and describes how they informed its design.

\subsection{CGEC datasets and error-tagged learner corpora}\label{existing-cgec-datasets}

Three benchmark resources have been especially influential, and each represents errors differently. NLPCC2018 \citep{zhao-etal-2018-nlpcc}, derived largely from Lang-8 learner journals, first segments the learner and corrected sentences into words. It then uses the MaxMatch ($M^2$) scorer \citep{dahlmeier-ng-2012-better} to align the two word sequences and group their differences into edits, such as missing, unnecessary, and replaced words. When several groupings are possible, the scorer selects the set of edits that best matches the reference correction. Because this process operates on words, different segmentation choices can change the number and extent of the resulting edits.
MuCGEC provides multiple reference corrections for many sentences and uses \texttt{ChERRANT} to extract character-based edit spans and labels automatically \citep{zhang-etal-2022-mucgec,hinson-etal-2020-heterogeneous}. The use of multiple references allows a learner sentence to have several possible repairs rather than a single fixed answer. YACLC likewise records alternative corrections and the number of annotators who proposed each one \citep{wang-etal-2021-yaclc}. It also distinguishes minimal grammatical correction from broader rewriting for fluency. This distinction parallels the contrast in learner-corpus research between minimal target hypotheses ($\text{TH}_1$), which correct strictly ungrammatical forms, and expanded or fluent target hypotheses ($\text{TH}_2$), which also improve idiomaticity and discourse naturalness \citep{ludeling-hirschmann-2015-error,reznicek-etal-2013-falko}. These benchmarks support correction and system evaluation.

Other Chinese learner corpora assign error categories manually. Examples include the HSK Dynamic Composition Corpus \citep{zhang-2009-features}, the Guangwai--Lancaster Chinese Learner Corpus \citep{chen-xu-2019-quantitative}, and the TOCFL learner corpus, whose labels combine broad edit operations with finer linguistic categories \citep{lee-etal-2018-building}. These categories provide useful linguistic information but were developed independently of the edit labels commonly used in CGEC. A layered representation can preserve both kinds of information while making the relationship between them explicit.

\subsection{Prior taxonomies of Chinese learner errors}

Prior classifications of Chinese learner errors range from broad pedagogical hierarchies to computationally oriented typologies. \citet{linlin-2006-error} provides one of the earliest systematic classifications. It organizes errors into character-, vocabulary-, and sentence-level categories, covering phenomena such as malapropisms (\zh{别字} \textit{biézì}), word-order errors, omission, redundancy, and mixed sentence patterns (\zh{句式杂糅} \textit{jùshì záróu}). This three-part organization is straightforward for instructional use, although it provides limited detail about the roles of particular word classes and sentence structures.

\citeauthor{dazhong-2020-analysis}'s (\citeyear{dazhong-2020-analysis}) monograph takes a more fine-grained pedagogical approach. Its thirty chapters each focus on a recurrent grammatical item or construction. Some chapters compare items that learners frequently confuse, such as \zh{才} \textit{cái} and \zh{就} \textit{jiù} or the three \textit{de} particles. Others explain how the parts of a construction work together. For example, distributive \zh{都} \textit{dōu} normally occurs with an expression referring to multiple members or a range, while the \zh{把} \textit{bǎ} construction requires an affected object and a predicate that indicates how that object is affected. In such cases, identifying the edited word alone does not fully explain the error; learners also need to understand the role of that word in the larger expression or construction. The monograph therefore offers detailed material for pedagogical diagnosis.

Finally, \citet{gu-etal-2025-improving} refine automatic Chinese ERRANT annotation with linguistically informed categories. Their implementation uses \texttt{stanza} part-of-speech tagging \citep{qi-etal-2020-stanza}. It labels sound-related substitutions as \texttt{R:PINYIN}, visually related substitutions as \texttt{R:SHAPE}, and substitutions involving both as \texttt{R:MULTI}. It also distinguishes character order (\texttt{R:CO}) from word order (\texttt{R:WO}) and assigns dedicated labels to the \textit{de} particles. These additions improve the description of automatically extracted edits, while pedagogical accounts provide complementary information about the learning difficulty or construction associated with an edit.

\subsection{Design procedure}\label{sec:design}

The taxonomy was developed by comparing three groups of sources: major CGEC datasets and tools \citep{zhao-etal-2018-nlpcc,zhang-etal-2022-mucgec,wang-etal-2021-yaclc,gu-etal-2025-improving}; pedagogically oriented classifications of Chinese learner errors \citep{linlin-2006-error,dazhong-2020-analysis}; and learner-corpus research on explicit coding, comparison across corpora, and the separation of observable edits from their interpretation \citep{dagneaux-etal-1998-computer,ludeling-hirschmann-2015-error}. The Falko German learner corpus provides a useful precedent for the last principle: it records the selected correction, or \textit{target hypothesis}, separately from the interpretation of the error \citep{reznicek-etal-2013-falko}.

For each source, we recorded the unit being annotated, the surface edit operation, the role of word segmentation, available part-of-speech information, the treatment of Chinese-specific phenomena, the distinction between minimal correction and fluency-oriented rewriting, and the amount of pedagogical interpretation supplied.

We then applied three tests to decide where each distinction belonged in the taxonomy. First, we retained a distinction if it could be determined from the learner sentence, its correction, and the immediate linguistic context; applied consistently across corpora; and used without relying on a particular textbook sequence or feedback style. Orthographic errors are annotated for edit operation and subtype; all other errors are annotated for edit operation, broad linguistic domain, and part of speech.

Second, a distinction was placed in the extension layer if it added useful diagnostic information for Chinese language teaching but was too fine-grained or context-dependent for cross-corpus comparison. {Examples include identifying the aspect marker \zh{了} \textit{le} with the extension \texttt{ASP:le}, the distributive adverb \zh{都} \textit{dōu} with \texttt{ADD:dou}, and the \zh{把} \textit{bǎ} construction with \texttt{CONST:ba}.}

Third, a distinction was excluded if assigning it would require unsupported assumptions about the internal cause of an error, the learner's unexpressed intention, or broader discourse-pragmatic meaning.

\section{A layered taxonomy for Chinese grammatical error annotation} \label{sec:taxonomy}

The basic unit of the taxonomy is an edit: a difference between a learner sentence and one selected correction. Annotators may identify edits manually or with an alignment tool. The position of an edit can be marked at the character level, without first segmenting the sentence into words.

Each annotation is relative to a \emph{target hypothesis}, the particular correction against which the learner sentence is compared \citep{reznicek-etal-2013-falko,ludeling-hirschmann-2015-error}. The same learner sentence may allow several repairs that produce different edits and labels. If a sentence has several accepted corrections, a project should designate one as primary or annotate each correction separately. Annotations based on different corrections are kept separate.

The taxonomy first separates errors in character form or writing conventions from other errors. For an orthographic error, the annotation records how the written form changes and the relation between the learner and corrected forms. For every other error, it records the edit operation, a broad linguistic domain, and the part of speech of the affected material. A separate optional layer can add a finer pedagogical diagnosis to non-orthographic errors.

This layered design separates the mechanics of correction from linguistic and pedagogical interpretation. It has three aims: to show exactly how the text changes, to connect labels with recognizable Mandarin grammatical categories, and to support meaningful feedback.

\subsection{Edit operations}

The operation records the observable change from the learner text to the selected correction, before any finer linguistic or pedagogical interpretation.

The full inventory comprises Missing (\texttt{M}), Replacement (\texttt{R}), Unnecessary (\texttt{U}), and Word Order (\texttt{WO}), plus an orthographic-only Character Order operation (\texttt{CO}). \texttt{M} is used when the correction inserts missing material, \texttt{R} when it exchanges one form for another, and \texttt{U} when it deletes redundant material. These three operations correspond to omission, substitution, and addition in many error annotation schemes.\footnote{Letter conventions differ across schemes: \texttt{ChERRANT} \citep{zhang-etal-2022-mucgec} uses \texttt{R} for \emph{redundant} and \texttt{S} for \emph{substitution}, corresponding to \texttt{U} and \texttt{R} here.} Word Order is treated as a distinct top-level operation rather than as a subtype of Replacement.

This choice is both conceptual and practical. In a word-order error the same material is preserved but appears in the wrong position. An automatic comparison may represent this movement as one deletion and one insertion, so \texttt{ChERRANT} applies an additional rule to identify some reordering spans \citep{hinson-etal-2020-heterogeneous,zhang-etal-2022-mucgec}. A span identified in this way receives \texttt{WO}, which records the underlying ordering problem directly. When a movement instead appears as separate deletion and insertion edits, each edit is labeled by the change it shows.\footnote{{An annotation project} may additionally link such paired edits, for example through a shared identifier and recorded source and target positions, to represent a long-distance movement explicitly.} \texttt{WO} marks word- and phrase-level reordering only; character transpositions within a word receive \texttt{CO}, adopted from \citet{gu-etal-2025-improving}, so that the same operation label never carries two senses. Orthographic \texttt{M} and \texttt{U} are reserved for missing and redundant punctuation, respectively.

\subsection{Orthographic screening and subtypes}
\label{sec:orth-subtypes}

Orthographic errors include wrong-character spellings, internal character reversals, and problems involving punctuation form or placement.

The scheme adapts five observable distinctions from \citet{gu-etal-2025-improving}. The first three classify character substitutions by the relation between the learner and corrected characters. \texttt{phon} marks a substitution between characters with identical or closely similar pronunciations but no salient visual similarity, as in \errcorr{做业}{zuòyè}{ill-formed spelling}{作业}{zuòyè}{homework}.\footnote{In examples throughout, an asterisk marks the learner form {as ungrammatical under the intended interpretation}.}

\texttt{shape} marks visually similar characters with different pronunciations, as in \errcorr{自已}{zì yǐ}{ill-formed spelling}{自己}{zìjǐ}{oneself}. \texttt{complex} is used when the characters are similar in both pronunciation and shape. An example is \errcorr{亚州}{Yàzhōu}{ill-formed spelling of Asia}{亚洲}{Yàzhōu}{Asia}.

\texttt{order} marks the correct characters in reversed order within one word, as in \errcorr{么什}{me shén}{ill-formed reversal}{什么}{shénme}{what}. Finally, \texttt{punc} marks the replacement, omission, or redundancy of a punctuation mark. The operation and subtype are recorded together as \textsc{op:orth:subtype}, where \textsc{op} represents the operation and \textsc{orth} marks the orthographic route. Table~\ref{tab:orth-labels} lists the seven operation--subtype combinations used in the scheme.

\begin{table}[!ht]
\singlespacing
\centering
\caption{Inventory of orthographic labels}
\label{tab:orth-labels}
\footnotesize
\renewcommand{\arraystretch}{1.15}
\begin{tabularx}{\textwidth}{>{\raggedright\arraybackslash}p{0.27\textwidth} Y}
\toprule
\textbf{Label} & \textbf{Use} \\
\midrule
\texttt{R:ORTH:phon} & Phonological identity or close similarity without salient visual similarity \\
\texttt{R:ORTH:shape} & Visual similarity with different pronunciation \\
\texttt{R:ORTH:complex} & Both phonological and visual similarity are salient \\
\texttt{CO:ORTH:order} & Correct characters in the wrong internal order within one word \\
\texttt{R:ORTH:punc} & Wrong punctuation form or choice \\
\texttt{M:ORTH:punc} & Missing punctuation mark \\
\texttt{U:ORTH:punc} & Redundant punctuation mark \\
\bottomrule
\end{tabularx}
\end{table}

The homophonous particles \zh{的} \textit{de}, \zh{地} \textit{de}, and \zh{得} \textit{de} remain non-orthographic because their correction concerns grammatical function. Their annotation therefore follows the non-orthographic route described next.

A correction between two established words is likewise non-orthographic when it changes lexical meaning. Thus, \zh{有益} \textit{yǒuyì} (`beneficial')~$\rightarrow$~\zh{有害} \textit{yǒuhài} (`harmful') is a content-word replacement.

\subsection{The non-orthographic core: linguistic domain and part of speech}

A non-orthographic edit receives one of three broad linguistic domains and a POS label. These layers distinguish content-word choice, functional-word choice, and phrasal or clausal structure.

\textbf{LEX-CONT (Lexical-Content)} errors involve choosing an inappropriate content word while the word's grammatical role remains stable. These include near-synonym confusions, collocation errors, register mismatches, and learner-created lexical items. For example, \zh{认识} \textit{rènshi} (`know a person; be acquainted with') and \zh{知道} \textit{zhīdào} (`know a fact') are both verbs, but they select different semantic objects. This domain covers open classes, the expandable vocabulary categories that include nouns, verbs, adjectives, and content adverbs.

\textbf{LEX-FUNC (Lexical-Functional)} errors involve choosing the wrong item from a relatively small set of grammatical words while the surrounding construction remains otherwise intact. Typical cases involve structural particles, aspect markers, negation markers, modal auxiliaries, classifiers, prepositions or coverbs, conjunctions, and adverbs such as \zh{都} \textit{dōu}. Linguists often call these \emph{closed classes} because they contain a relatively restricted set of members. The key diagnostic is that the learner has selected the wrong item from a restricted functional inventory, not that the clause pattern itself must be rebuilt.

\textbf{STRUCT (Structural)} errors concern phrase and clause structure: whether a required element is present, correctly positioned, and correctly combined with its neighbors. This domain covers argument structure, complement formation, clause-level word order, the placement and interpretation of words such as negatives and quantifiers, and constructional patterns such as the \zh{把} \textit{bǎ}-construction, \zh{被} \textit{bèi}-passive, \zh{比} \textit{bǐ}-comparative, serial-verb patterns, and \zh{连$\cdots$也/都$\cdots$} \textit{lián$\cdots$yě/dōu$\cdots$} focus construction. Structural labels are used when the correction changes whether an element is permitted or required, where it appears, how subjects and objects are expressed, or how parts of a construction combine. This domain also includes mixed patterns such as combining incompatible \zh{把} \textit{bǎ} and \zh{被} \textit{bèi} structures.

The POS inventory for this route is VERB (verb), NOUN (noun), ADJ (adjective), ADV (adverb), PART (grammatical particle), AUX (auxiliary verb), PREP (preposition), CLF (classifier or measure word), and CONJ (coordinating or subordinating conjunction). The inventory can be mapped to existing Chinese tagsets or to Universal Dependencies categories, depending on the corpus.\footnote{The disposal marker \zh{把} \textit{bǎ} and the passive marker \zh{被} \textit{bèi} are treated here as coverbs and tagged \texttt{PREP}, following standard descriptive and pedagogical analyses of Mandarin \citep{chao-1968-grammar,li-thompson-1981-mandarin}; \citet{dazhong-2020-analysis} likewise classifies \zh{被} \textit{bèi} as a preposition (\zh{介词}), while noting an alternative particle analysis motivated by the optional omission of its object. 
{Annotation projects that adopt construction-specific tags} may instead map these markers to dedicated treebank categories or to the Universal Dependencies \texttt{ADP}/\texttt{AUX} labels.} \texttt{X} is reserved for spans where no single tag applies. Projects that need finer resolution may extend the inventory, for example by adding dedicated tags for pronouns, determiners, and numerals,\footnote{In the proposed taxonomy, these categories receive the closest inventory tag.} provided that the mapping is documented with the annotations.

Automatic tools such as \texttt{stanza} \citep{qi-etal-2020-stanza} or LTP \citep{che-etal-2021-n} can suggest POS labels. Because word segmentation and POS tagging can be unreliable in learner data, annotators verify the suggested segmentation and tags before assigning a POS label.

\subsection{Decision rules and interaction among layers}\label{sec:interactions}

Each non-orthographic edit receives a composite label following the template \textsc{op:dom:pos}, where \textsc{op} denotes the edit operation, \textsc{dom} the linguistic domain, and \textsc{pos} the part of speech associated with the edited item, span, or construction.

Domain assignment follows a fixed rule based on the operation and the affected grammatical material. A Replacement involving a content word is \texttt{LEX-CONT}, because the correction substitutes one lexical choice for another. A Replacement involving an item from a relatively small set of grammatical words is \texttt{LEX-FUNC}: for example, replacing \zh{的} \textit{de} with \zh{地} \textit{de} in an adverbial modifier is a functional-word choice. Every non-orthographic Missing, Unnecessary, or Word Order edit is assigned to \texttt{STRUCT}: omitted \zh{把} \textit{bǎ}, redundant \zh{被} \textit{bèi}, and misordered \zh{都} \textit{dōu} are all structural because the correction changes whether a required element is present or correctly positioned. Constructional confusions that surface as replacements, such as exchanging \zh{把} \textit{bǎ} for \zh{被} \textit{bèi}, remain \texttt{LEX-FUNC} at the core level; the constructional diagnosis is carried by the extension layer (e.g., \texttt{MIX:ba-bei}).

Replacement is treated differently from Missing, Unnecessary, and Word Order because it changes the item occupying an existing position, whereas the other operations change whether material is present or correctly ordered. The latter therefore receive \texttt{STRUCT}, even when the affected item is a function word such as the aspect marker \zh{了} \textit{le} or the structural particle \zh{的} \textit{de}. The extension layer preserves the affected item's identity: \texttt{M:STRUCT:PART} with extension \texttt{ASP:le} reads as a structural omission of a missing \zh{了} \textit{le}. This fixed division also avoids subjective judgments about whether a construction is otherwise stable.

Under this convention, the operation and POS layers usually determine the domain. The domain layer is nonetheless retained explicitly: it makes labels self-documenting for users who do not work with POS tags, it records a genuine decision for adverbs, and it allows projects to adopt the operation and domain layers without committing to a POS inventory. An explicit list guides the remaining decision for adverbs. An adverb counts as functional when it belongs to one of the listed classes for temporal sequencing, additivity and distributivity, contrast, negation, or degree (Table~\ref{tab:extension-domains}); other adverbs are treated as content words.

\begin{figure}[htbp]
\centering
\caption{{Annotation flow. The orthographic route yields a label following the template \textsc{op:orth:subtype}. The non-orthographic route yields a label following \textsc{op:dom:pos} and may be followed by a pedagogical extension.}}
\label{fig:layer-interaction}
\resizebox{\textwidth}{!}{%
\begin{tikzpicture}[
    node distance=9mm and 20mm,
    decisionbox/.style={
        draw,
        diamond,
        aspect=2.5,
        align=center,
        inner sep=4pt
    },
    startbox/.style={
        draw,
        rounded corners,
        align=center,
        inner sep=6pt,
        minimum width=0.50\linewidth
    },
    orthbox/.style={
        draw,
        rounded corners,
        align=center,
        inner sep=6pt,
        minimum width=0.43\linewidth
    },
    widebox/.style={
        draw,
        rounded corners,
        align=center,
        inner sep=6pt,
        minimum width=0.78\linewidth
    },
    rulebox/.style={
        draw,
        rounded corners,
        align=center,
        inner sep=6pt,
        minimum width=0.90\linewidth
    },
    resultbox/.style={
        draw,
        rounded corners,
        align=center,
        inner sep=6pt,
        minimum width=0.62\linewidth
    },
    optionalbox/.style={
        draw,
        dashed,
        rounded corners,
        align=center,
        inner sep=6pt,
        minimum width=0.62\linewidth
    },
    arrow/.style={->, thick}
]

\node[startbox] (edit) {Edit defined relative to a selected target hypothesis};

\node[decisionbox, below=of edit] (check) {ORTH?};

\node[orthbox, right=28mm of check] (orth) {
    \textbf{Orthographic label}\\[2pt]
{\textsc{op:orth:subtype}}\\
{\scriptsize \texttt{phon, shape, complex, order, punc}}
};

\node[widebox, below=17mm of check] (operation) {
    \textbf{Determine non-orthographic operation}\\[2pt]
    \texttt{M \quad R \quad U \quad WO}
};

\node[widebox, below=of operation] (pos) {
    \textbf{Assign POS to the affected material}\\[2pt]
    \texttt{VERB \quad NOUN \quad PART \quad ADV \quad ... \quad X}
};

\node[rulebox, below=of pos] (rule) {
    \textbf{Deterministic domain rule}\\[2pt]
    Every non-orthographic \texttt{M}, \texttt{U}, or \texttt{WO}
    $\rightarrow$ \texttt{STRUCT}\\
    \texttt{R} + content item $\rightarrow$ \texttt{LEX-CONT};
    \qquad
    \texttt{R} + closed-class or functionally restricted item
    $\rightarrow$ \texttt{LEX-FUNC}
};

\node[resultbox, below=of rule] (core) {
    \textbf{Non-orthographic core}\\[2pt]
{\textsc{op:dom:pos}}
};

\node[optionalbox, below=of core] (extension) {
    \textbf{Optional pedagogical extension}
};

\draw[arrow] (edit) -- (check);
\draw[arrow] (check) -- node[above]{Yes} (orth);
\draw[arrow] (check) -- node[left]{No} (operation);
\draw[arrow] (operation) -- (pos);
\draw[arrow] (pos) -- (rule);
\draw[arrow] (rule) -- (core);
\draw[arrow] (core) -- node[right]{if applicable} (extension);

\end{tikzpicture}}
\end{figure}
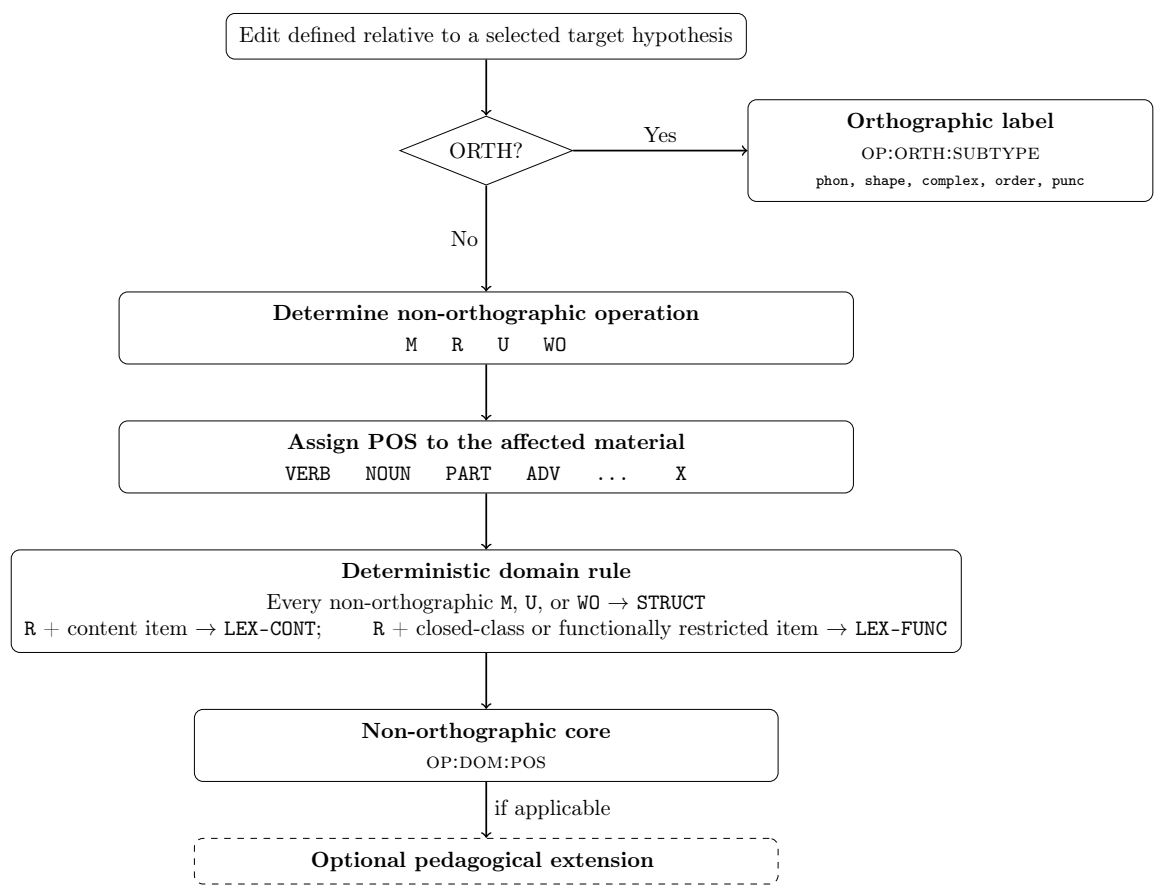

Figure~\ref{fig:layer-interaction} summarizes the annotation flow; Section~\ref{sec:extensions} describes optional extensions for non-orthographic edits.

\section{Pedagogical extensions}\label{sec:extensions}

Non-orthographic core labels provide a shared basis for corpus comparison and system evaluation. {A pedagogical extension may accompany a core label following the template \textsc{op:dom:pos} to identify the functional item or construction involved.}

Projects may adopt only the modules relevant to their teaching or research aims, but each adopted module should be applied consistently. Each extension is recorded separately and combines a family code and an item, separated by a colon. For example, \texttt{M:STRUCT:PART} shows that a grammatical particle is missing, while \texttt{ASP:le} identifies it as the aspect marker \zh{了} \textit{le}.

\subsection{Extension inventory}

Many word- and syntax-level errors in Chinese learner writing concern the function of a grammatical item or the form of a construction. The optional functional--constructional codes record this information. Word-level codes group small sets of items that share discourse, semantic, or grammatical functions; constructional codes identify families such as comparison, argument structure, and complements. The inventory builds on \citet{dazhong-2020-analysis} and adds a family when it represents a clearly defined set of items or constructions that recurs in learner data. For example, \texttt{MOD} covers modal auxiliaries, an established area of Chinese pedagogical grammar \citep{li-thompson-1981-mandarin}. The \texttt{MIX} code represents blends of incompatible constructions described in pedagogical taxonomies such as \citet{linlin-2006-error}.

Word-level codes cover temporal sequencing (\texttt{TEMP}), additivity and distributivity (\texttt{ADD}), contrastive stance (\texttt{CONT}), coordination and clause linking (\texttt{LINK}), locatives and path (\texttt{LOC}), aspect (\texttt{ASP}), and modality (\texttt{MOD}). Constructional codes cover focus (\texttt{FOC}), argument structure and voice (\texttt{CONST}), the expression and placement of arguments (\texttt{ARG}), comparison and degree (\texttt{DEG}), and complement families (\texttt{COMP}).

Table~\ref{tab:extension-domains} summarizes these extension families and also lists the structural-particle (\texttt{STR}), negation (\texttt{NEG}), and classifier (\texttt{CLF}) codes used to refine functional-word errors.

\begin{table}[!ht]
\singlespacing
\centering
\caption{Extension codes for fine-grained pedagogical annotation, adapted and extended from \citet{dazhong-2020-analysis}.}
\label{tab:extension-domains}
\scriptsize
\renewcommand{\arraystretch}{1.2}
\resizebox{\textwidth}{!}{%
\begin{tabular}{p{0.31\textwidth} p{0.11\textwidth} p{0.56\textwidth}}
\hline
\textbf{Extension family} & \textbf{Code} & \textbf{Representative items / constructions} \\
\hline
Temporal sequencing and evaluation & TEMP & \zh{才} \textit{cái} (`only then'), \zh{就} \textit{jiù} (`then; already'), \zh{再} \textit{zài} (`again; next') \\
Additivity and distributivity & ADD & \zh{也} \textit{yě} (`also'), \zh{都} \textit{dōu} (`all'), \zh{凡是} \textit{fánshì} (`whenever; all that') \\
Contrastive stance & CONT & \zh{却} \textit{què} (`yet'), \zh{反而} \textit{fǎn'ér} (`on the contrary') \\
Coordination and clause linking & LINK & \zh{和} \textit{hé} (`and'), \zh{或者} \textit{huòzhě} (`or'), \zh{还是} \textit{háishi} (`or' in questions), \zh{不管} \textit{bùguǎn}/\zh{无论} \textit{wúlùn} (`regardless of') \\
Degree and comparison & DEG & \zh{真} \textit{zhēn} (`really'), \zh{一点儿} \textit{yìdiǎnr} (`a little'), \zh{比} \textit{bǐ} (`than') \\
Locatives and path expressions & LOC & \zh{从} \textit{cóng} (`from'), \zh{在} \textit{zài} (`at; in'), \zh{到} \textit{dào} (`to; arrive') \\
Aspect and event development & ASP & \zh{着} \textit{zhe}, \zh{了} \textit{le}, \zh{过} \textit{guo}, \zh{动词重叠} \textit{dòngcí chóngdié} (`verb reduplication') \\
Modality and volition & MOD & \zh{能} \textit{néng} (`can; be able'), \zh{会} \textit{huì} (`can; know how'), \zh{可以} \textit{kěyǐ} (`may; be allowed'), \zh{应该} \textit{yīnggāi} (`should'), \zh{必须} \textit{bìxū} (`must'), \zh{要} \textit{yào} and \zh{想} \textit{xiǎng} in auxiliary use \\
Focus and emphasis & FOC & \zh{连$\cdots$也/都$\cdots$} \textit{lián$\cdots$yě/dōu$\cdots$} (`even ... also/all ...') \\
Argument-structure and voice constructions & CONST & \zh{使} \textit{shǐ} causatives, \zh{把} \textit{bǎ} disposal constructions, \zh{被} \textit{bèi} passives; the code names the construction (e.g., \texttt{CONST:ba}) \\
Argument realization and placement & ARG & misplaced or doubled arguments within a clause or construction; the code names the grammatical role (e.g., \texttt{ARG:object}) \\
Complements & COMP & \zh{结果补语} \textit{jiéguǒ bǔyǔ} (`result complement'), \zh{时量补语} \textit{shíliàng bǔyǔ} (`duration complement'), \zh{趋向补语} \textit{qūxiàng bǔyǔ} (`directional complement'), \zh{可能补语} \textit{kěnéng bǔyǔ} (`potential complement'), \zh{状态/程度补语} \textit{zhuàngtài/chéngdù bǔyǔ} (`state/degree complement') \\
Structural particles & STR & \zh{的} \textit{de}, \zh{地} \textit{de}, \zh{得} \textit{de} (attributive, adverbial, and complement particles) \\
Negation & NEG & \zh{不} \textit{bù} (general negation), \zh{没} \textit{méi} (perfective/existential negation) \\
Classifiers & CLF & \zh{个} \textit{gè}, \zh{本} \textit{běn}, \zh{张} \textit{zhāng}, \zh{条} \textit{tiáo}, and other classifier--noun pairings \\
Mixed construction patterns & MIX & \zh{句式杂糅} \textit{jùshì záróu} (`mixed sentence patterns'), especially incompatible combinations of constructional frames \\
\hline
\end{tabular}}
\end{table}

\subsection{Assignment and scope}

When a code names a word, the item is written in lowercase pinyin. For a Replacement, it names the contrasting pair in a fixed order, as in \texttt{TEMP:cai-jiu}, \texttt{ASP:le-guo}, and \texttt{NEG:bu-mei}; the edit itself shows the direction of change. Missing, Unnecessary, and Word Order normally name a single item, as in \texttt{ASP:le} or \texttt{ADD:dou}. The code \texttt{STR:de} covers the three structural particles \zh{的}, \zh{地}, and \zh{得}.

Construction-based extensions name a construction, subtype, or grammatical role, as in \texttt{CONST:ba}, \texttt{COMP:result}, and \texttt{ARG:object}. A missing \zh{都} \textit{dōu} in the \zh{连$\cdots$都} \textit{lián$\cdots$dōu} construction therefore receives \texttt{FOC:lian-dou} rather than \texttt{ADD:dou}.

When a named construction licenses or organizes the edited material, its construction-level code takes precedence over an item-level code. Otherwise, the item-level code is used.

The extension layer is limited to closed-class items and recurring constructions with relatively small or clearly defined inventories.

Open-class lexical errors are more difficult to divide consistently. A near-synonym choice, an unusual collocation, and a learner-created form may overlap, and their interpretation depends on open-ended knowledge of usage. The core therefore records such errors with labels such as \texttt{R:LEX-CONT:VERB}, without requiring a lexical subtype. A project may add local subtypes when its research question requires them, provided that they remain optional and are documented separately from the shared core.

\section{Reference specification of the proposed taxonomy}\label{sec:comprehensive}

Table~\ref{tab:orth-labels} {in Section~\ref{sec:taxonomy}} presents the orthographic labels. Table~\ref{tab:error-taxonomy} presents the non-orthographic core and optional pedagogical extensions, with a definition and example for each. The tables map the character-level categories of \citet{linlin-2006-error} and \citet{gu-etal-2025-improving} and the functional--constructional analyses of \citet{dazhong-2020-analysis} onto the proposed taxonomy.
The examples are illustrative rather than exhaustive. When a correction depends on the intended meaning or discourse context, the English gloss states that interpretation explicitly.\footnote{Gloss abbreviations (ATTR `attributive particle'; LE, BA, BEI for the markers \zh{了} \textit{le}, \zh{把} \textit{bǎ}, \zh{被} \textit{bèi}; CLF `classifier'; Q `question marker') follow the spirit of the Leipzig Glossing Rules. They appear only in learner-side glosses and only where an idiomatic translation would obscure the error.}
In the Extension column, codes are drawn from the functional--constructional, structural-particle, negation, and classifier families listed in Table~\ref{tab:extension-domains} and follow the naming conventions described in Section~\ref{sec:extensions}.\footnote{There is no extension for lexical-content errors; the Extension cell is marked with an em dash (---).}
{The POS tag} \texttt{X} marks a non-orthographic span for which no single category applies, including whole constructions and multi-word spans without one main word.

{\scriptsize
\singlespacing
\setlength{\tabcolsep}{2pt}
\renewcommand{\arraystretch}{1.18}
\begin{longtable}{
  >{\raggedright\arraybackslash}p{0.045\textwidth}
  >{\raggedright\arraybackslash}p{0.065\textwidth}
  >{\raggedright\arraybackslash}p{0.145\textwidth}
  >{\raggedright\arraybackslash}p{0.285\textwidth}
  >{\raggedright\arraybackslash}p{0.350\textwidth}
}
\caption{Reference specification of the non-orthographic portion of the proposed taxonomy, organized by operation, domain, part of speech, and optional pedagogical extension}
\label{tab:error-taxonomy} \\

\toprule
\textbf{Op} &
\textbf{POS} &
\textbf{Optional extension} &
\textbf{Definition / Pattern} &
\textbf{Example} \\
\midrule
\endfirsthead

\toprule
\textbf{Op} &
\textbf{POS} &
\textbf{Optional extension} &
\textbf{Definition / Pattern} &
\textbf{Example} \\
\midrule
\endhead

\midrule
\multicolumn{5}{r}{Continued on next page} \\
\endfoot
\bottomrule
\endlastfoot

\multicolumn{5}{>{\raggedright\arraybackslash}p{0.86\textwidth}}{\textbf{LEX-CONT (Lexical-Content Errors; Replacement of an Open-Class Content Word)}} \\

R & VERB & --- &
Near-synonym confusion between content verbs &
\errcorr{我知道他，我们是好朋友}{wǒ zhīdào tā, wǒmen shì hǎo péngyou}{I know of him; we are good friends}{我认识他，我们是好朋友}{wǒ rènshi tā, wǒmen shì hǎo péngyou}{I know him; we are good friends} \\

R & NOUN & --- &
Near-synonym confusion between content nouns &
\errcorr{一次难忘的经验}{yí cì nánwàng de jīngyàn}{an unforgettable experience (know-how)}{一次难忘的经历}{yí cì nánwàng de jīnglì}{an unforgettable experience (event)} \\

R & VERB & --- &
Collocation error in a verb--object combination &
\errcorr{做决心}{zuò juéxīn}{make determination}{下决心}{xià juéxīn}{make up one's mind} \\

\midrule
\multicolumn{5}{>{\raggedright\arraybackslash}p{0.86\textwidth}}{\textbf{LEX-FUNC (Lexical-Functional Errors; Function-Word Replacement in an Otherwise Unchanged Construction)}} \\

R & PART & STR:de &
Wrong choice among structural particles \zh{的}/\zh{地}/\zh{得} \textit{de} &
\errcorr{慢慢的走}{mànmàn de zǒu}{walk slow-ATTR}{慢慢地走}{mànmàn de zǒu}{walk slowly} \\

R & ADV & NEG:bu-mei &
Wrong choice between negation markers \zh{不} \textit{bù} and \zh{没} \textit{méi} &
\errcorr{昨天我不去}{zuótiān wǒ bù qù}{Yesterday I do not go}{昨天我没去}{zuótiān wǒ méi qù}{Yesterday I did not go} \\

R & PART & ASP:le-guo &
Wrong choice among aspect markers &
\errcorr{我以前看了这个电影}{wǒ yǐqián kàn le zhège diànyǐng}{I watched-LE this movie before}{我以前看过这个电影}{wǒ yǐqián kànguo zhège diànyǐng}{I have seen this movie before} \\

R & AUX & MOD:neng-hui &
Wrong modal auxiliary (general ability vs.\ acquired skill) &
\errcorr{他能说法语}{tā néng shuō Fǎyǔ}{He can (is able to) speak French}{他会说法语}{tā huì shuō Fǎyǔ}{He can (has learned to) speak French} \\

R & ADV & TEMP:cai-jiu &
Wrong temporal sequencing marker &
\errcorr{他十点就来}{tā shí diǎn jiù lái}{He came as early as ten}{他十点才来}{tā shí diǎn cái lái}{He did not come until ten} \\

R & ADV & ADD:ye-dou &
Wrong additive or distributive marker &
\errcorr{这三个学生也来了}{zhè sān ge xuéshēng yě lái le}{These three students also came}{这三个学生都来了}{zhè sān ge xuéshēng dōu lái le}{These three students all came} \\

R & PREP & LOC:cong-dao &
Wrong location or path marker while the surrounding clause pattern is otherwise unchanged &
\errcorr{我从学校去}{wǒ cóng xuéxiào qù}{I go from school}{我到学校去}{wǒ dào xuéxiào qù}{I go to school} \\

R & CLF & CLF:ge-ben &
Wrong classifier for the main noun (the head noun) &
\errcorr{一个书}{yí ge shū}{one CLF book}{一本书}{yì běn shū}{one book} \\

R & ADV & CONT:que-faner &
Wrong contrastive-stance adverb (simple contrast vs.\ counter-expectation) &
\errcorr{他不但没生气，却笑了}{tā búdàn méi shēngqì, què xiào le}{He not only did not get angry, yet smiled}{他不但没生气，反而笑了}{tā búdàn méi shēngqì, fǎn'ér xiào le}{Not only did he not get angry; on the contrary, he smiled} \\

R & CONJ & LINK:huozhe-haishi &
Choice marker mismatched with clause type (interrogative \zh{还是} \textit{háishi} in a declarative) &
\errcorr{我想去北京还是上海}{wǒ xiǎng qù Běijīng háishi Shànghǎi}{I want to go to Beijing or-Q Shanghai}{我想去北京或者上海}{wǒ xiǎng qù Běijīng huòzhě Shànghǎi}{I want to go to Beijing or Shanghai} \\

\midrule
\multicolumn{5}{>{\raggedright\arraybackslash}p{0.86\textwidth}}{\textbf{STRUCT (Structural Errors)}} \\

M & PREP & CONST:ba &
Missing marker required by a \zh{把} \textit{bǎ}-construction &
\errcorr{我书看完了}{wǒ shū kàn wán le}{I book finished reading}{我把书看完了}{wǒ bǎ shū kànwán le}{I finished reading the book} \\

U & PREP & CONST:bei &
Unnecessary passive marker \zh{被} \textit{bèi} with an intransitive verb &
\errcorr{这件事被发生了}{zhè jiàn shì bèi fāshēng le}{This matter was happened}{这件事发生了}{zhè jiàn shì fāshēng le}{This matter happened} \\

M & PART & ASP:le &
Missing aspect marker required by the event meaning or a surrounding adverb &
\errcorr{我已经吃饭}{wǒ yǐjīng chī fàn}{I already eat}{我已经吃了饭}{wǒ yǐjīng chī le fàn}{I have already eaten} \\

U & PART & ASP:le &
Redundant aspect marker &
\errcorr{昨天我吃了饭了}{zuótiān wǒ chī le fàn le}{Yesterday I ate LE meal LE}{昨天我吃了饭}{zuótiān wǒ chī le fàn}{Yesterday I ate a meal} \\

M & PART & ASP:zhe &
Missing \zh{着} \textit{zhe} marking an ongoing state in an existence--location pattern &
\errcorr{墙上挂一张画}{qiáng shàng guà yì zhāng huà}{On the wall hangs a picture}{墙上挂着一张画}{qiáng shàng guàzhe yì zhāng huà}{A picture is hanging on the wall} \\

M & VERB & COMP:result &
Missing resultative complement required by the meaning of the construction &
\errcorr{请把作业写}{qǐng bǎ zuòyè xiě}{Please BA homework write}{请把作业写完}{qǐng bǎ zuòyè xiěwán}{Please finish writing the homework} \\

M & PART & COMP:degree &
Missing complement marker \zh{得} \textit{de} in degree construction &
\errcorr{他跑快}{tā pǎo kuài}{He run fast}{他跑得快}{tā pǎo de kuài}{He runs fast} \\

M & ADV & FOC:lian-dou &
Missing \zh{都}/\zh{也} \textit{dōu}/\textit{yě} required by the \zh{连$\cdots$都/也} \textit{lián$\cdots$dōu/yě} focus construction &
\errcorr{他连一个字没写}{tā lián yí ge zì méi xiě}{He even one character did not write}{他连一个字都没写}{tā lián yí ge zì dōu méi xiě}{He did not write even a single character} \\

U & ADV & DEG:bi &
Redundant degree adverb inside a \zh{比} \textit{bǐ} comparative &
\errcorr{他比我很高}{tā bǐ wǒ hěn gāo}{He than me very tall}{他比我高}{tā bǐ wǒ gāo}{He is taller than I am} \\

WO & ADV & ADD:dou &
Adverb whose meaning depends on its position placed incorrectly &
\errcorr{都他们去}{dōu tāmen qù}{All they go}{他们都去}{tāmen dōu qù}{They all go} \\

WO & X & LOC:zai &
Time or location phrase in an incorrect position (misplaced \zh{在} \textit{zài}-phrase) &
\errcorr{我在学校昨天学习}{wǒ zài xuéxiào zuótiān xuéxí}{I at school yesterday studied}{我昨天在学校学习}{wǒ zuótiān zài xuéxiào xuéxí}{I studied at school yesterday} \\

WO & NOUN & ARG:object &
Object placed in an incorrect structural position &
\errcorr{我把看书完了}{wǒ bǎ kàn shū wán le}{I BA read book finish}{我把书看完了}{wǒ bǎ shū kànwán le}{I finished reading the book} \\

WO & ADV & COMP:potential &
Negation placed outside the potential-complement pattern &
\errcorr{我不听懂}{wǒ bù tīngdǒng}{I not hear-understand}{我听不懂}{wǒ tīng bu dǒng}{I cannot understand (what I hear)} \\

U & PREP & MIX:ba-bei &
Mixing incompatible syntactic patterns (redundant \zh{被} \textit{bèi} inside a \zh{把} \textit{bǎ}-construction) &
\errcorr{我把书被看完了}{wǒ bǎ shū bèi kànwán le}{I BA book BEI finished reading}{我把书看完了}{wǒ bǎ shū kànwán le}{I finished reading the book} \\

\end{longtable}
}

\section{Evaluation of the taxonomy}
\label{sec:adequacy}

To evaluate the coverage and consistency of the proposed taxonomy, we conducted two analyses. The first counts how often the dedicated categories occur in the MuCGEC development set. The second asks five LLMs to label a sample of these edits and measures their agreement. Both analyses use edits produced by \citeauthor{gu-etal-2025-improving}'s (\citeyear{gu-etal-2025-improving}) linguistically informed extension of \texttt{ChERRANT}. Its output can be reused directly. The four operations correspond to those of the proposed scheme, the sound- and shape-based categories supply the orthographic subtype, and the POS tag carries over once the edit span is matched. Domain and extension labels are assigned separately (Supplementary Table S16).

\subsection{Category coverage in MuCGEC development data}\label{sec:coverage}
To examine how often the dedicated categories occur in CGEC data, we analyzed edit records extracted from the first correction supplied for each sentence in the MuCGEC development set. Of 1,137 sentences, 58 were excluded because the first ``correction'' was an annotator note such as \zh{没有错误} \textit{méiyǒu cuòwù} (`no error') or \zh{无法标注} \textit{wúfǎ biāozhù} (`cannot annotate'). The tool produced 4,545 edit records for the remaining sentences. We removed 119 records that did not represent learner errors: 116 contained no character change, and three inserted an annotator note left in the reference text. The analysis therefore covers 4,426 edits. Pooling the edits from all alternative references yields closely similar proportions.

Table~\ref{tab:coverage} gives the full distribution. Replacements account for 55.9\% of the edits, Missing for 27.4\%, and Unnecessary for 16.7\%. Orthographic screening identifies 24.0\% of all edits, divided almost equally between character substitutions based on sound or visual similarity and punctuation. The automatic tool marks very few reorderings. This reflects the difficulty of detecting moved material automatically; some movements instead appear as paired deletions and insertions.

A cautious automatic count based on predefined function words places another 17.6\% of edits in the extension inventory. It includes only single-word edits whose source or correction contains a listed item, so it misses multi-word constructional repairs and words whose form is shared with a common content word. Among the remaining edits, about half involve open-class words, a quarter involve broader or mixed-category rewordings, and the rest include function words not yet covered by the list. Modal auxiliaries formed a clear group in this last set and motivated the addition of \texttt{MOD}; other candidates include role-marking prepositions and additional linking expressions.

Two patterns support specific design choices. Among the 185 edits that the tool assigns a dedicated \textit{de}-particle label, 168 are insertions or deletions and only 17 are replacements; the \texttt{STR} trigger count of 227 also includes \textit{de} edits that the tool labels in other ways. This distribution accords with treating missing or unnecessary particles as Structural errors and particle choice as Lexical-Functional. Replacements involving an adverb, where an annotator must choose between a content and functional analysis, form only 1.7\% of all edits.

These figures describe category coverage for this automatically extracted set of edits; other word-segmentation and alignment procedures can produce different edit units.

\begin{table}[!ht]
\singlespacing
\centering
\caption{Distribution of the 4,426 first-reference MuCGEC development edits, after excluding 58 sentences whose references were annotator comments and 119 tool-produced records without learner errors. Categories below the operation panel are mutually exclusive.}
\label{tab:coverage}
\scriptsize
\begin{tabular}{l >{\raggedright\arraybackslash}p{0.42\textwidth} r r}
\toprule
\textbf{Group} & \textbf{Category} & \textbf{Edits} & \textbf{\%} \\
\midrule
Edit operation & Replacement (\texttt{R}) & 2,472 & 55.9 \\
 & Missing (\texttt{M}) & 1,213 & 27.4 \\
 & Unnecessary (\texttt{U}) & 741 & 16.7 \\
\midrule
Orthographic screening & similarity-typed substitution (\texttt{R:PINYIN}/\texttt{SHAPE}/\texttt{MULTI}) & 526 & 11.9 \\
 & punctuation & 536 & 12.1 \\
 & character order (\texttt{R:CO} $\rightarrow$ \texttt{CO:ORTH:order}) & 2 & $<$0.1 \\
 & \emph{subtotal} & 1,064 & 24.0 \\
\midrule
Flagged word-order reordering & \texttt{R:WO} & 9 & 0.2 \\
\midrule
Extension triggers & \texttt{STR} (\zh{的}/\zh{地}/\zh{得}) & 227 & 5.1 \\
(single-token, & \texttt{MOD} & 99 & 2.2 \\
non-orthographic) & \texttt{LOC} & 91 & 2.1 \\
 & \texttt{ASP} & 81 & 1.8 \\
 & \texttt{CLF} & 63 & 1.4 \\
 & \texttt{ADD} & 62 & 1.4 \\
 & \texttt{TEMP}, \texttt{LINK}, \texttt{NEG}, \texttt{CONST}, \texttt{DEG}, \texttt{CONT} & 155 & 3.5 \\
 & \emph{subtotal} & 778 & 17.6 \\
\midrule
Remaining & other edits & 2,575 & 58.2 \\
\bottomrule
\end{tabular}
\end{table}

\subsection{A preliminary test of annotation consistency with LLMs}\label{sec:llmprobe}

The written specification should lead annotators to similar decisions. {To examine this expectation, we asked five LLMs to apply the same taxonomy to the same edits.} Comparing their decisions shows which parts of the instructions they interpret similarly and where category boundaries remain unclear.

\paragraph{Setup}
{We tested five LLMs from five providers on 391 items.\footnote{Before records without a character change were removed, we drew a simple random sample of 400 from the 4,545 automatically extracted records. Nine sampled records contained no character change and were excluded, leaving 391 items.} Each item included the learner sentence, its correction, the marked edit, and the annotation guideline. The models labeled the items independently, and three were run twice to assess within-model consistency. Supplementary Section S5 reports the model identifiers, settings, instructions, response-processing procedure, and additional analyses.}

The models first decided whether an edit was orthographic and then applied the layers relevant to that route: operation and subtype for orthographic edits; operation, domain, POS, and extension for all others. Raw agreement records how often two labels match; Fleiss' $\kappa$ and Krippendorff's $\alpha$ additionally adjust for agreement expected by chance \citep{artstein-poesio-2008-survey}. Table~\ref{tab:llm-agreement} reports these measures separately for each layer. A partial-credit comparison also shows how often some, but not all, layers match.

{One of the authors independently assigned reference labels} to a deliberately varied subset of 115 items: 40 on which the models were unanimous, 39 with a majority decision, and 36 on which they were divided.\footnote{Before records without a character change were removed, we selected 40 items from each group. Five of the 120 contained no character change and were excluded, leaving 115 items.} The author reviewed them without seeing the model identities or responses. When more than one analysis was supported by the written guideline, the reference record retained acceptable alternatives. The final table column compares the most common model label with this expert reference, averaging over tied labels. Because difficult, divided cases are overrepresented, the column describes this subset rather than the full sample.

\begin{table}[!ht]
\singlespacing
\centering
\caption{Agreement among five LLMs applying the taxonomy to 391 MuCGEC edits. For layers beyond orthographic screening, an item contributes to a pairwise comparison only when both models assign that layer. Composite agreement combines the layers applicable to each route. Extensions are compared at the broad category level. Denominators in the final column vary by layer within the 115-item expert-reviewed subset.}
\label{tab:llm-agreement}
\scriptsize
\setlength{\tabcolsep}{5pt}
\renewcommand{\arraystretch}{1.2}
\begin{tabular}{l r r r r}
\toprule
\textbf{Layer} & \textbf{Pairwise (\%)} & \textbf{$\kappa$} & \textbf{$\alpha$} & \textbf{Panel label vs.\ expert (\%)} \\
\midrule
Orthographic screening decision & 97.1 & 0.883 & 0.883 & 99.1 \\
Orthographic label (operation and subtype) & 92.7 & 0.898 & 0.898 & 100.0 \\
Non-orthographic edit operation & 91.6 & 0.878 & 0.878 & 85.6 \\
Linguistic domain & 89.8 & 0.823 & 0.823 & 82.7 \\
Part of speech & 84.6 & 0.820 & 0.820 & 76.1 \\
Extension code & 84.5 & 0.736 & 0.737 & 75.7 \\
Composite label (exact, code level) & 69.9 & 0.688 & 0.689 & 53.8 \\
Composite (partial credit) & 86.1 & -- & -- & -- \\
\bottomrule
\end{tabular}
\end{table}

Three patterns are most relevant to taxonomy development. First, orthographic screening and non-orthographic edit operation show the strongest agreement, while the optional extension has the lowest agreement after adjustment for chance; raw agreement for part of speech and the extension is nearly identical. Exact agreement on the whole composite label is lower, but rises from 69.9\% to 86.1\% when matching layers receive partial credit. The models also follow the fixed relationship between edit operations and the Structural domain in nearly all applicable cases.

Second, the repeated runs of the same model match on approximately four fifths of the composite labels. The remaining within-model variation provides useful context for interpreting the lower agreement between different models.

Third, disagreements cluster around a few interpretable category distinctions. Models differ between Replacement and Missing when a correction adds material alongside an unchanged character, between lexical-content and lexical-functional analyses for some adverbs and rewordings, over when a multi-word span requires the POS label \texttt{X}, and over whether a pedagogical extension applies. Agreement with the expert reference follows the same pattern: unanimous model labels match the expert much more often than labels from divided cases. These disagreements identify places where the guidelines need clearer examples and decision rules. Reliability with trained human annotators remains to be examined.

\section{Discussion}\label{sec:discussion}

The assessment suggests that the taxonomy's layers serve different but complementary purposes. By design, every edit receives either an orthographic label or a non-orthographic core label, and the MuCGEC analysis indicates that the dedicated orthographic and functional--constructional categories are relevant to a substantial part of the data. The consistency study indicates that surface operation labels are easier to apply than optional pedagogical interpretations. These findings support compact shared labels with optional pedagogical detail where a project has the evidence, training, and research need to use it.

For learner-corpus annotation, the layered format makes the relation between a correction and its interpretation visible. For an orthographic edit, the operation and subtype describe the change in written form; for a non-orthographic edit, the operation records what changed, the domain and POS identify the broad linguistic area, and an extension can name a particular Chinese item or construction. These layers also keep the annotation tied to a target hypothesis. When two accepted corrections analyze a learner sentence differently, their labels remain attached to their respective source--correction pairs rather than being combined into one apparently certain diagnosis. Compatibility with ChERRANT offers a practical route for existing CGEC resources: projects can reuse automatically derived spans and operations, then add linguistic or pedagogical information appropriate to their aims.

The pedagogical extensions provide hypotheses about distinctions that may help teachers and learners. Labels for aspect, structural particles, comparison, argument structure, and complements can support profiles organized around recognizable topics of Chinese grammar. Their value, however, depends on use: studies with teachers and learners should examine whether the additional labels improve interpretation, feedback, curriculum planning, or learner uptake.

Human, automatic, and LLM-assisted annotation can each use the taxonomy at a different stage. Automatic alignment can propose edit spans and surface operations; an LLM can suggest labels or explanations for review; and trained annotators can decide among context-dependent analyses and make the final annotation decision. This division of work reflects recent learner-corpus studies that use LLM suggestions with human review and Chinese GEC research that uses LLMs for explanations and evaluation \citep{gajo-etal-2025-learn,acharya-etal-2025-tracing,li-etal-2025-rethinking}. The present comparison remains preliminary because model versions can change, different models may share training data or failure patterns \citep{pangakis-etal-2023-automated}, and the expert reference was produced by one author. A fuller reliability study should train several human annotators, record agreement separately for each layer, and use a separate reviewer to resolve disagreements and revise the guidelines.

Several aspects of coverage also require broader testing. The extension inventory draws substantially on one pedagogical source, and the coverage analysis uses one dataset and one automatic method to prepare the data. Further studies should compare corpora representing different proficiency levels, learner language backgrounds, genres, and correction policies. They should also compare methods for word segmentation and edit matching. Such work can determine which extensions recur across settings and which are best kept specific to individual projects.

The present scheme concentrates on errors that can be identified through a direct comparison between a learner sentence and its correction. Discourse-level problems involving coherence, reference tracking, and information structure require wider context and may need additional annotation layers. Extending the scheme to such phenomena would carry a broader learner-corpus principle---recording observed changes separately from their interpretation---from sentence-level annotation toward a fuller account of Chinese learner writing.

\section{Conclusion}\label{sec:conclusion}

{This paper has developed a layered taxonomy that connects edit-based CGEC annotation with linguistic and pedagogical analysis of Chinese learner errors. Rather than combining surface change and pedagogical interpretation in a single category, the taxonomy uses a two-route architecture. Orthographic errors follow the template \textsc{op:orth:subtype}, whereas non-orthographic errors receive a core label following \textsc{op:dom:pos} and may receive an optional functional or constructional extension. Because every annotation remains tied to a selected target hypothesis, alternative accepted corrections can be represented without collapsing their potentially different analyses.}

{The assessment provides initial evidence for the usefulness of this separation. In the 4,426 MuCGEC edits examined, orthographic screening identified 24.0\% of the data, and a cautious automatic procedure found extension triggers in another 17.6\%. The preliminary study of five LLMs on 391 items found the greatest consistency for orthographic screening and surface edit operations, with lower agreement as part-of-speech and pedagogical distinctions were combined into complete labels. These findings suggest that the compact layers offer a comparatively stable basis for shared annotation, while the more interpretive categories require clearer guidance and further validation.}

{The taxonomy is intended to be modular: existing \texttt{ChERRANT} spans and operations can be reused, while linguistic domains, POS tags, and pedagogical extensions can be added according to the aims of a corpus or instructional setting. Future work should evaluate reliability among trained human annotators, test coverage across learner populations, proficiency levels, genres, and correction policies, and examine whether the extensions improve feedback and pedagogical interpretation. Additional layers may also be needed for discourse-level phenomena that cannot be diagnosed from a sentence--correction pair alone. By keeping observable edits separate from their linguistic and pedagogical interpretation, the proposed framework provides a transparent basis for learner-corpus comparison, instructional analysis, and human-reviewed automatic or LLM-assisted annotation.}

\section*{Declaration of generative AI in the manuscript preparation process}

During the preparation of this work, the authors used OpenAI ChatGPT and Codex to assist with language editing, sentence restructuring, manuscript organization, and improving the clarity and consistency of the technical exposition. The authors reviewed and edited all AI-assisted content, verified the formal arguments, and take full responsibility for the content of the article.


\end{document}